\documentclass[sigconf]{acmart}

\usepackage{amsmath}
\usepackage{amsfonts}
\usepackage{bm}
\usepackage{multirow}
\usepackage{subfigure}
\AtBeginDocument{%
  \providecommand\BibTeX{{%
    \normalfont B\kern-0.5em{\scshape i\kern-0.25em b}\kern-0.8em\TeX}}}

\setcopyright{acmcopyright}
\copyrightyear{2021}
\acmYear{2021}
\acmDOI{10.1145/1122445.1122456}

\acmConference[CIKM '21]{The 30th ACM International Conference on Information and Knowledge Management (CIKM '21)}{November 01--05, 2021}{Queensland, Australia}
\acmBooktitle{CIKM '21: International Conference on Information and Knowledge Management, November 01--05, 2021, Queensland, Australia}
\acmPrice{15.00}
\acmISBN{978-1-4503-XXXX-X/18/06}

\begin{document}

\title{Lifelong Intent Detection via Multi-Strategy Rebalancing}

\author{Qingbin Liu, Xiaoyan Yu, Shizhu He, Kang Liu, Jun Zhao}
\email{qingbin.liu,xiaoyan.yu,shizhu.he,kliu,jzhao@nlpr.ia.ac.cn}
\authornotemark[1]
\affiliation{%
  \institution{National Laboratory of Pattern Recognition, Institute of Automation, Chinese Academy of Sciences}
  \postcode{100190}
}
\affiliation{%
  \institution{School of Artificial Intelligence, University of Chinese Academy of Sciences}
  \postcode{100049}
}

\begin{abstract}
Conventional Intent Detection (ID) models are usually trained offline, which relies on a fixed dataset and a predefined set of intent classes. However, in real-world applications, online systems usually involve continually emerging new user intents, which pose a great challenge to the offline training paradigm. Recently, lifelong learning has received increasing attention and is considered to be the most promising solution to this challenge. In this paper, we propose Lifelong Intent Detection (LID), which continually trains an ID model on new data to learn newly emerging intents while avoiding catastrophically forgetting old data. Nevertheless, we find that existing lifelong learning methods usually suffer from a serious imbalance between old and new data in the LID task. Therefore, we propose a novel lifelong learning method, Multi-Strategy Rebalancing (MSR), which consists of cosine normalization, hierarchical knowledge distillation, and inter-class margin loss to alleviate the multiple negative effects of the imbalance problem. Experimental results demonstrate the effectiveness of our method, which significantly outperforms previous state-of-the-art lifelong learning methods on the ATIS, SNIPS, HWU64, and CLINC150 benchmarks.
\end{abstract}

\ccsdesc[500]{Computing methodologies~Online learning settings}
\ccsdesc[500]{Computing methodologies~Discourse, dialogue and pragmatics}

\keywords{Lifelong Learning, Intent Detection, Multi-Strategy Rebalancing}


\maketitle

\section{Introduction}
Intent Detection (ID) aims to accurately understand the user intent from a user utterance to guide downstream dialogue policy decisions \cite{ATIS, SNIPS, UID}. It is an essential component of dialogue systems and is therefore widely used in real-world applications, such as personal assistants and customer service. In these systems, ID models usually classify a user utterance into an intent class. For example, an ID model should be able to recognize the intent of ``booking a flight'' from the utterance ``I am flying to Chicago next Wednesday''.

\begin{figure}[h]
  \centering
  \includegraphics[width=2.3in]{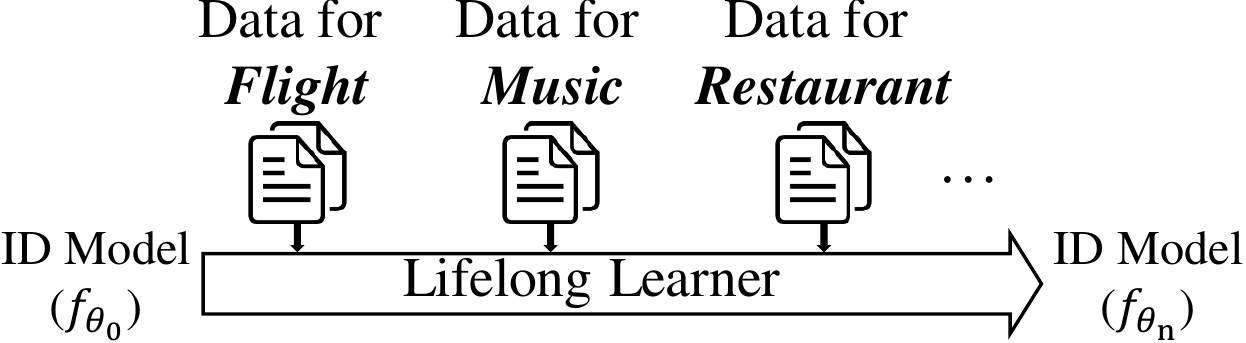}
  \caption{Lifelong Intent Detection: The lifelong learning method (Lifelong Learner) continually trains an ID model when new data becomes available.}
  \Description{Lifelong Intent Detection: The lifelong learning method (Lifelong Learner) continually trains an ID model when new data becomes available.}
  \label{fig1}
\end{figure}

Existing ID models usually adopt an offline learning paradigm, which performs once-and-for-all training on a fixed dataset. This paradigm can only handle a fixed number of user intents. However, online dialogue systems typically need to handle continually emerging new user intents, which makes previous ID models impractical in real-world applications. Recently, lifelong learning has received increasing attention and is considered to be the most promising approach to address this problem \cite{LL1, LL2}. Therefore, to handle continually emerging new intents, we propose the Lifelong Intent Detection (LID) task, which introduces lifelong learning into the ID task. As shown in Fig \ref{fig1}, the LID task continually trains an ID model using only new data to learn newly emerging intents. At any time, the updated ID model should be able to perform accurate classifications for all classes observed so far. In this task, it is infeasible to retrain the ID model from scratch every time new data becomes available due to storage budgets and computational costs \cite{KCN}.

A plain lifelong learning method is to fine-tune a model pre-trained on old data directly on new data. However, this method faces a serious challenge, namely catastrophic forgetting, where models fine-tuned on new data usually suffer from a significant performance degradation on old data \cite{mccloskey1989catastrophic,french1999catastrophic}. To address this issue, the current mainstream lifelong learning methods either identify and retain parameters that are important to the old data \cite{ewc,aljundi2018memory}, or maintain a memory to reserve a small number of old training samples (known as the reply-based methods) \cite{icarl,emr}. At each time, reply-based methods combine the reserved old data with the new data to retrain the model. Due to the simplicity and effectiveness of replay-based methods, they become an excellent solution for lifelong learning in natural language processing scenarios \cite{emar,KCN}.
\begin{figure*}
  \centering
  \includegraphics[width=4.4in]{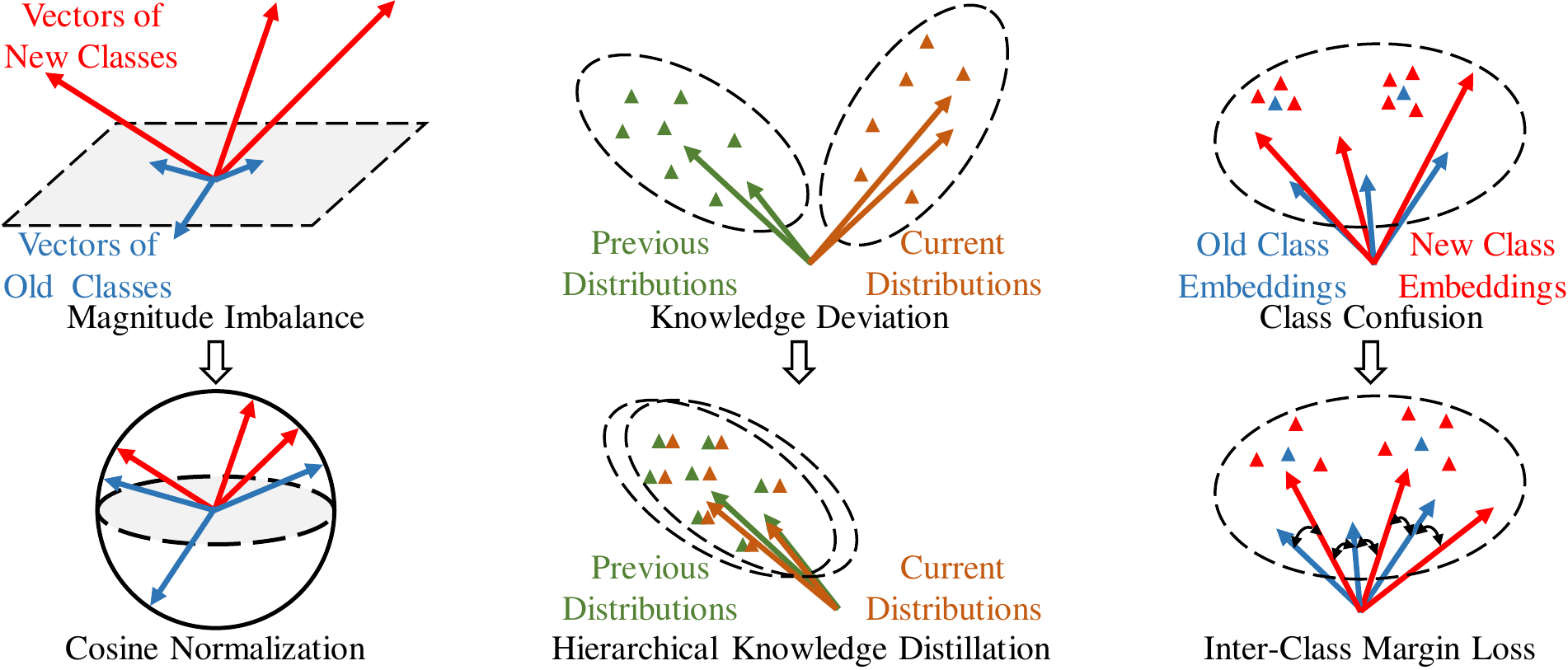}
  \caption{Illustrations of the multiple negative effects caused by the data imbalance problem in the LID task and our solutions.}
  \label{fig2}
  \Description{The adverse effects caused by the data imbalance and our solutions.}
\end{figure*}

However, when adapting existing replay-based methods to lifelong intention detection, our study found that these methods suffer from a data imbalance problem. Specifically, at each step of the lifelong learning process, there is generally a large amount of new class data, yet only a small amount of old data is reserved, leading to a significant imbalance between old and new data. Under such circumstances, the focus of the training process will be significantly biased towards new classes, thus leading to a series of negative effects in the ID model, as shown in Figure \ref{fig2}: (1) Magnitude Imbalance: the magnitude of feature vectors and class embeddings of new classes is significantly larger than those of old classes, (2) Knowledge Deviation: the knowledge of the previous model, i.e., the feature distribution and the probability distribution of old classes, is not well preserved, (3) Class Confusion: the class embeddings of new classes and those of old classes are very close to each other in the high-dimensional vector space. These adverse effects severely mislead the ID model, causing it to tend to predict new classes while catastrophically forgetting old classes.

Our work is inspired by lifelong learning in image classification tasks \cite{NCM, EEIL, FSCIL}, which also targets the data imbalance problem. In this paper, we find multiple adverse effects caused by the imbalance problem in the LID task and propose corresponding solutions.

To address the problem of data imbalance, we propose a novel lifelong learning framework, namely Multi-Strategy Rebalancing (MSR), which aims to learn a balanced ID model. Specifically, MSR contains three components to alleviate the above three adverse effects: (1) Cosine Normalization, which balances the magnitude of feature vectors and class embeddings between old and new classes by constraining these vectors in a high-dimensional sphere to eliminate the bias caused by the difference in magnitude. (2) Hierarchical Knowledge Distillation, which preserves the knowledge of the previous model from the feature level and the prediction level to retain the feature distribution and the probability distribution of old classes. (3) Inter-Class Margin Loss, which provides a large margin to separate the new class embeddings and the old class embeddings. With multi-strategy rebalancing, the ID model can effectively handle the adverse effects caused by data imbalance. We constructed four benchmarks for the LID task based on four widely used ID datasets to systematically compare different lifelong learning methods \cite{ATIS, SNIPS, HWU64, CLINC150}. Experimental results show that our proposed framework significantly outperforms previous state-of-the-art lifelong learning methods on these benchmarks.

In summary, the contributions of this work are as follows: 
\begin{itemize}
    \item To the best of our knowledge, we are the first to propose the Lifelong Intent Detection task, meanwhile constructed four benchmarks through four widely used ID datasets: ATIS, SNIPS, HWU64, and CLINC150.
    \item We propose the Multi-Strategy Rebalancing framework, which can effectively handle the data imbalance problem in the LID task through cosine normalization, hierarchical knowledge distillation, and inter-class margin loss. 
    \item Experimental results show that our method outperforms previous lifelong learning methods and achieves state-of-the-art performance. The source code and benchmarks will be released for further research (\url{http://anonymous}).
\end{itemize}

\section{Task Formulation}
Intent detection is usually formulated as a multi-class classification task, which predicts an intent class for a given user utterance \cite{ATIS,SNIPS,zhang-etal-2019-joint,e-etal-2019-novel}. In real-world applications, online systems inevitably face continually emerging new user intents. Therefore, we propose the Lifelong Intent Detection task, which continually trains the ID model on emerging data to learn new classes. In this task, there is a sequence of $K$ data $(D_1,D_2,... ,D_K)$. Each data ($D_i$) has its own label set ($C_i$), i.e., one or more intent classes, and training/validation/testing sets ($D^{\text{train}}_i$, $D^{\text{valid}}_i$, $D^{\text{test}}_i$). At each step, the lifelong learning framework trains the ID model on the new training set ($D^{\text{train}}_{i}$) to learn the new classes in $C_i$. The LID task requires that the ID model should perform well on all observed classes. Therefore, after training on $D^{\text{train}}_i$, the updated ID model will be evaluated on all observed testing sets (i.e., $\tilde{D}_k^\text{test} = \bigcup_{k=1}^{i}D_k^\text{test}$) and uniformly classify each sample into all known classes (i.e., $\tilde{C}_i = \bigcup_{k=1}^{i}C_i$).
\section{Method}
\begin{figure*}
  \centering
  \includegraphics[width=4.7in]{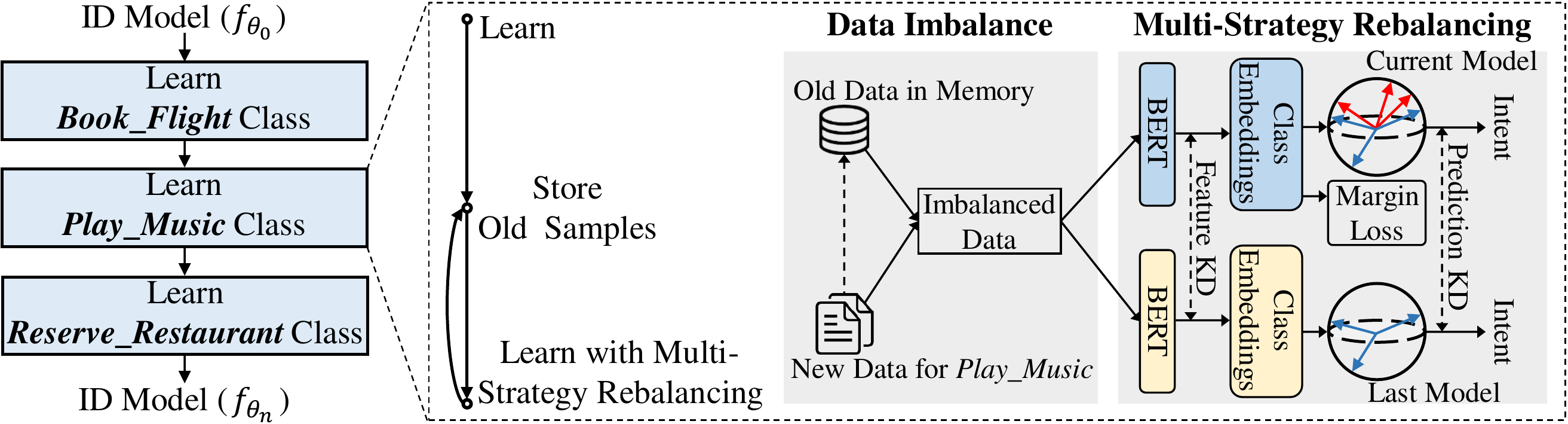}
  \caption{Illustrations of our method for lifelong intent detection. At each step, our method combines Cosine Normalization, Hierarchical Knowledge Distillation (KD), and Inter-Class Margin Loss to learn the imbalanced data.}
  \label{fig3}
  \Description{Illustrations of our method for lifelong intent detection. At each step, our method combines Cosine Normalization, Hierarchical Knowledge Distillation (KD), and Inter-Class Margin Loss to learn the imbalanced data.}
\end{figure*}

In this work, we propose Multi-Strategy Rebalancing to handle the data imbalance problem in the LID task. In this section, we will first show a typical replay-based method, iCaRL \cite{icarl}, as the background. Next, we deeply analyze the data imbalance problem and introduce the proposed solutions, which are shown in Figure \ref{fig3}.
\subsection{Background}
A typical ID model contains two components: an encoder and multiple class embeddings. The encoder can be recurrent neural networks or pre-trained models \cite{gru, BERT}. We adopt the current best encoder, BERT \cite{BERT}, as our encoder. BERT is a multi-layer Transformer \cite{transformers} that is pre-trained on large-scale unlabeled corpora. It encodes each sample into a sentence-level feature vector, i.e., the hidden state of the ``[CLS]'' token. Then, the ID model calculates the dot product similarity between the feature vector and the class embeddings as the class probability. The loss of the ID model is the standard cross-entropy loss:
\begin{equation}
\mathcal{L}_{\text{ce}}(x) = - \sum \nolimits^{|\tilde{C}|}_{i = 1} \bm{y}_i \text{log}(\bm{p}_i),
\end{equation}
where $\tilde{C}$ is the set of all observed classes. $\bm{y}$ is the one-hot ground-truth label. $\bm{p}$ is the class probability obtained by softmax.

To overcome catastrophically forgetting old data, iCaRL \cite{icarl} maintains a bounded memory to store a few representative old samples, which aims to introduce important information about the data distribution of previous classes into the training process. The memory can be denoted as $M$, where $M_i$ is the set of samples reserved for the $i-$th class. After training on the new data, iCaRL selects the most representative samples for each class in this data through a class prototype \cite{snell2017prototypical}, which is calculated by averaging the feature vectors of all training samples of that class. Based on the distance between the feature vector of each training sample and the prototype, iCaRL sorts the training samples of each class and selects the top $B/t$ nearest samples as exemplars to store, where $B$ is the memory size and $t$ is the number of all observed classes. To allocate space for the current classes, iCaRL removes $B/(t-m) - B/t$ training samples for each old class, where $m$ is the number of new classes. iCaRL removes samples that are far from the prototype according to the sorted list. In this way, the most representative samples are reserved in the memory.

In addition, iCaRL combines the cross-entropy loss with a knowledge distillation (KD) loss \cite{hinton2015distilling} to retrain the model. The distillation loss enables the model at the current step to learn the probability distribution of the model trained in the last step: 
\begin{equation}
\label{eq2}
\mathcal{L}_\text{kd}(x) = - \sum \nolimits^{|C^\text{o}|}_{i = 1} \gamma_i(\bm{s}^*)\text{log}({\gamma}_i(\bm{s}))
\end{equation}
where $\bm{s}^*$ and $\bm{s}$ are the soft labels (i.e., the results before the softmax layer) predicted by the last model and the current model for old classes ($C^\text{o}$), respectively. $\gamma_i(\bm{s}) =  {e^{\bm{s}_i/T}} / {\sum_{j=1}^{|C^\text{o}|} e^{\bm{s}_j/T}}$. $T$ is the temperature scalar, which is used to increase the weight of small probability values. The KD loss is an effective way to alleviate catastrophic forgetting by learning the soft label of the last model.

However, at each step, the new data is usually significantly more than the reserved old data, leading to a serious data imbalance problem. It makes previous methods tend to predict new classes and catastrophically forgetting old classes.

\subsection{Multi-Strategy Rebalancing}
In this work, we address the data imbalance problem from multiple aspects by incorporating three components, cosine normalization, hierarchical knowledge distillation, and inter-class margin loss. 
\subsubsection{Cosine Normalization}
We find that the magnitude of both feature vectors and class embeddings of new classes is significantly larger than that of old classes. It may make the current model tend to predict new classes. To solve this problem, we replace the original dot product similarity with cosine normalization as:
\begin{equation}
\bm{p}_i{(x)} = \frac {\exp(\tau \langle f(x), \theta_i \rangle)} {\sum_j^{|\tilde{C}|} \exp(\tau \langle f(x), \theta_j \rangle)}
\end{equation}
where $\langle f(x), \theta_i \rangle$ measures the cosine similarity between the feature vector $f(x)$ and the class embedding $\theta_i$. The hyper-parameter $\tau$ is used to control the peak of the softmax distribution since the cosine similarity ranges between -1 and 1. Geometrically, we constrain these vector in a high-dimensional sphere to effectively eliminate the bias caused by the imbalanced magnitudes.

\subsubsection{Hierarchical Knowledge Distillation}
The knowledge (i.e., the feature distribution and the probability distribution) of the model trained on new data usually deviates heavily from that of the model trained on old data. It makes the model forget the important information of old classes. We propose hierarchical knowledge distillation to preserve the previous knowledge from two levels.

In the \textbf{Feature-Level} KD, we reserve the geometric structure of the feature vector of the current model by reducing the angle between it and the feature vector of the last model:
\begin{equation}
\mathcal{L}_\text{fkd}(x)  = 1 - \langle {f}(x), {f}^{*}(x) \rangle
\end{equation}
where ${f}^{*}(x)$ is the feature vector extracted by the last model. $\mathcal{L}_\text{fkd}(x)$ encourages the features extracted by the current model to be close to the features extracted by the last model in the high-dimensional sphere. Besides, we fix the old class embeddings to reserve their spatial structure.

\begin{figure*}[htbp]
	\centering
	\subfigure[ATIS Benchmark]{
		\begin{minipage}[t]{0.2425\linewidth}
			\centering
			\includegraphics[width=1.74in]{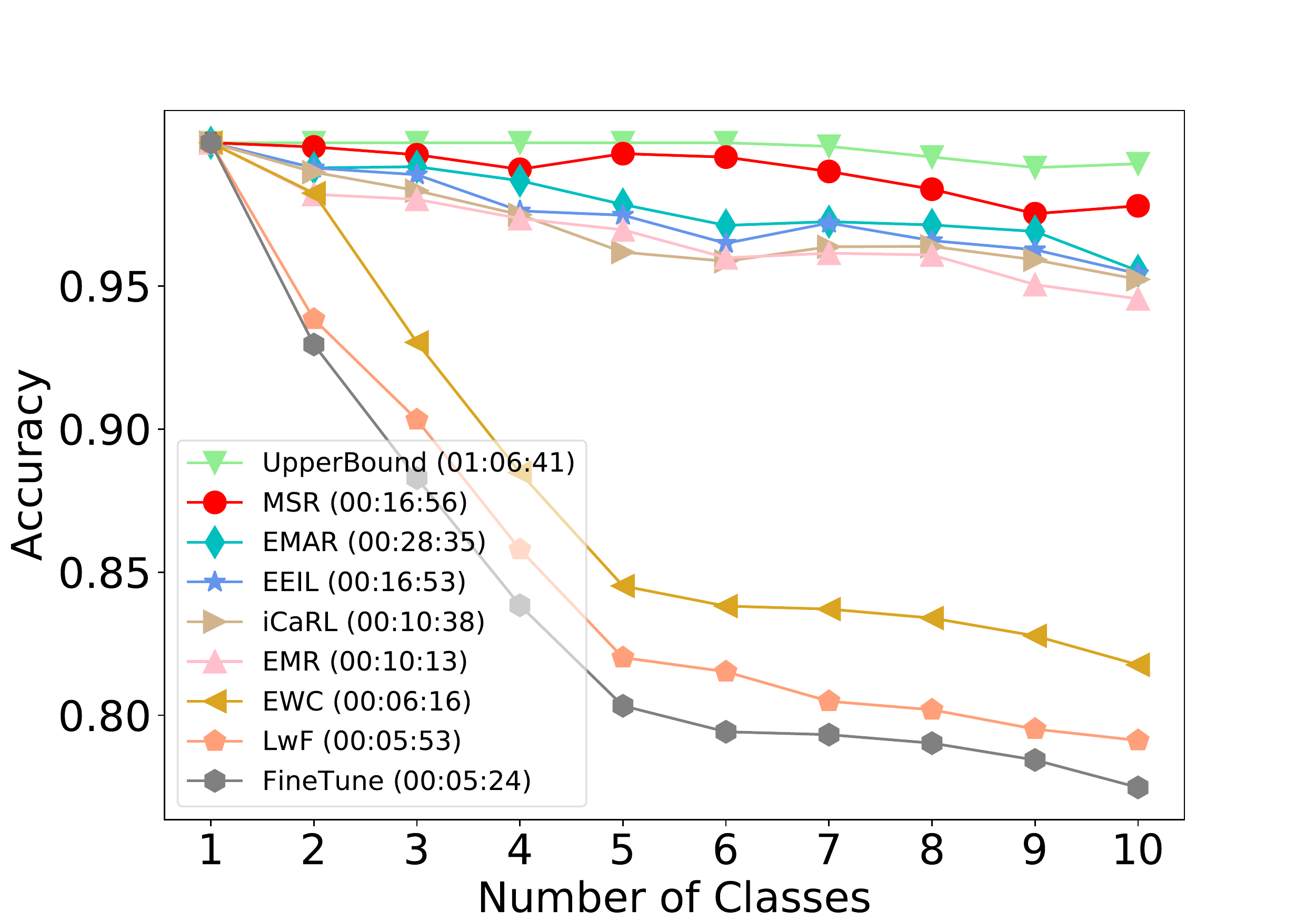}
		\end{minipage}%
	}
	\subfigure[SNIPS benchmark]{
		\begin{minipage}[t]{0.2425\linewidth}
			\centering
			\includegraphics[width=1.74in]{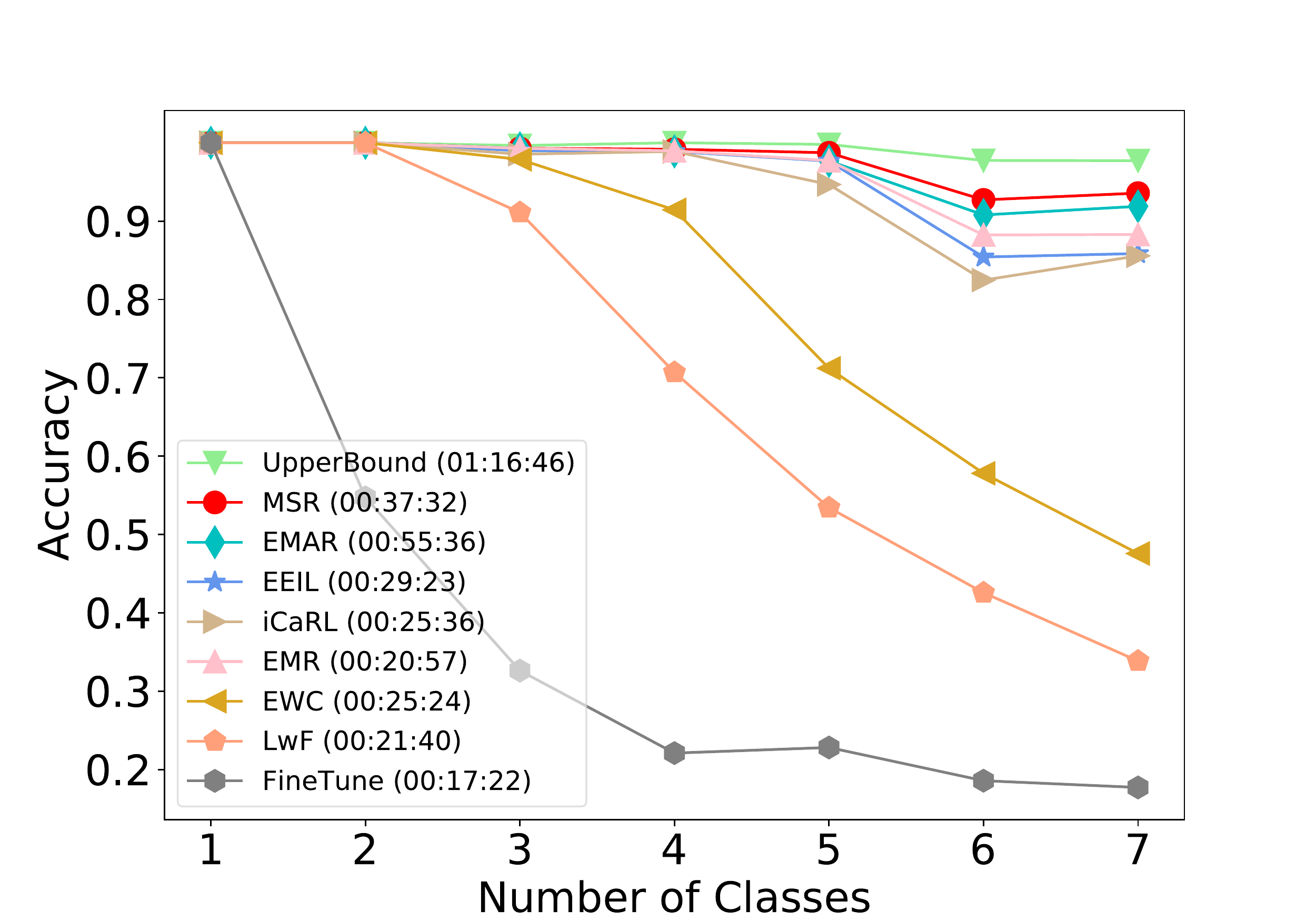}
		\end{minipage}%
	}
	\subfigure[HWU64 benchmark]{
		\begin{minipage}[t]{0.2425\linewidth}
			\centering
			\includegraphics[width=1.74in]{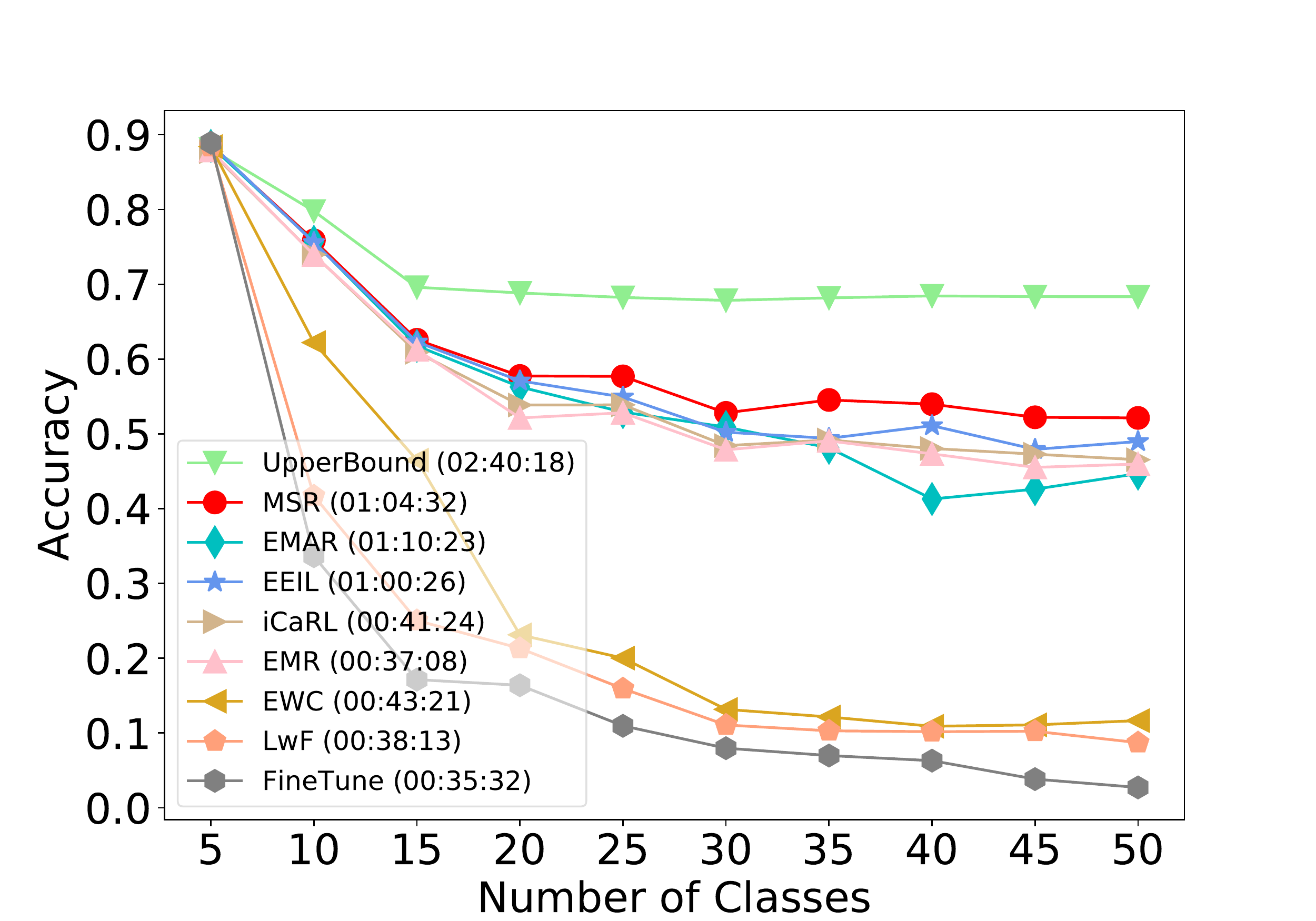}
		\end{minipage}%
	}
	\subfigure[CLINC150 benchmark]{
		\begin{minipage}[t]{0.2425\linewidth}
			\centering
			\includegraphics[width=1.74in]{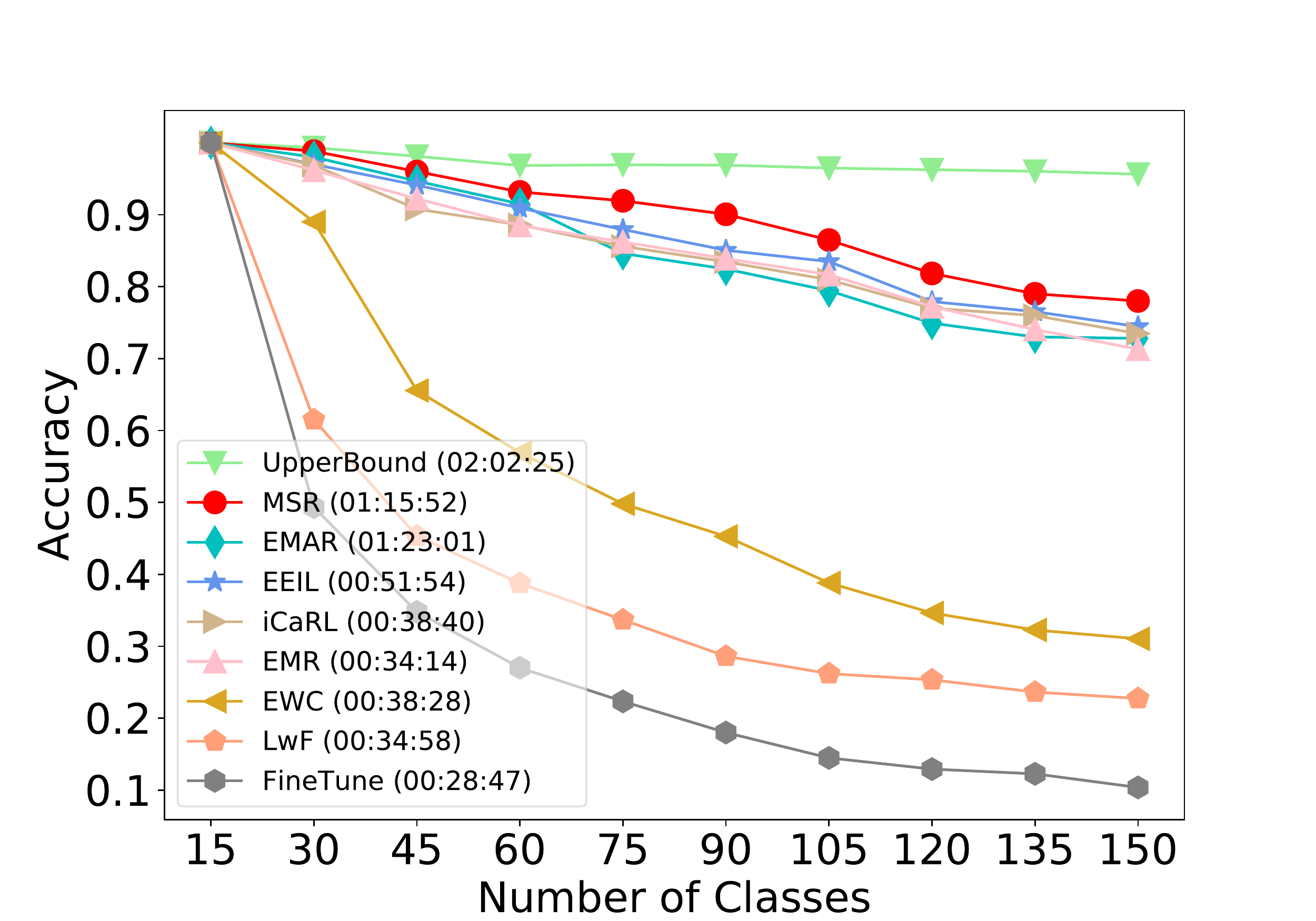}
		\end{minipage}%
	}
	\centering
	\caption{Performance ($\text{acc}_i$) changes with increasing classes on the ATIS, SNIPS, HWU64, CLINC150 benchmarks, respectively. We show the \textbf{training time} (measured on GeForce RTX 2080Ti) in the brackets.}
	\label{fig4}
\end{figure*}
In the \textbf{Prediction-Level} KD, we encourage the current model to reserve the probability distribution of the last model through a knowledge distillation loss, as in Eq. \ref{eq2}, which learns the soft label predicted by the last model. 
\subsubsection{Inter-Class Margin Loss}
Another negative effect of the imbalance problem is class confusion, i.e., new and old class embeddings are usually mixed in the high-dimensional space. This is due to the fact that a large number of new training samples are likely to activate neighboring samples with different labels \cite{NCM, FSCIL}. To solve this problem, we introduce an inter-class margin loss to separate these class embeddings as:
\begin{equation}
\mathcal{L}_\text{ICML}(\theta)  = \sum \nolimits_i^{|C|} \sum \nolimits_j^{|C^\text{o}|} \text{max}(\langle \theta_i, \theta_j \rangle - \alpha, 0)
\end{equation}
where $\alpha$ is the margin. This loss expects the angle between ($\theta_i$,$\theta_j$) to be greater than $\alpha$. Through this loss, these embeddings can be uniformly distributed on the high-dimensional sphere without confusion.
\subsection{Training}
At each step of LID, our MSR framework combines the above losses to train the ID model on the new data and the reserved old data. The overall loss is defined as follows:
\begin{equation}
\mathcal{L}  = \mathcal{L}_{\text{ce}}(x) + \beta_1 \mathcal{L}_{\text{kd}}(x) + \beta_2 \mathcal{L}_{\text{fkd}}(x) + \beta_3 \mathcal{L}_{\text{ICML}}(\theta)
\end{equation}
where $\beta_1$, $\beta_2$, and $\beta_3$ are hyper-parameters to balance the performance between old and new classes. $\mathcal{L}_{\text{ce}}$, $\mathcal{L}_{\text{kd}}$, and $\mathcal{L}_{\text{fkd}}$ are calculated for both the new data and the reserved old data. $\mathcal{L}_{\text{ICML}}$ is calculated for all new class embeddings.

\section{Experiment}
\subsection{Lifelong Intent Detection Benchmarks}
Since we are the first to propose the LID task, we construct four benchmarks based on the following method: for an ID dataset, we arrange its classes in a fixed random order. Each class has its own data. In a class-incremental manner, the lifelong learning methods continually train an ID model on one or multiple new classes. Based on four widely used datasets, ATIS \cite{ATIS}, SNIPS \cite{SNIPS}, HWU64 \cite{HWU64}, CLINC150 \cite{CLINC150}, we constructed four benchmarks. To provide a comprehensive evaluation, we set different numbers of new classes per step in different benchmarks. We set 1, 1, 5, and 15 new classes per step in the ATIS, SNIPS, HWU64, and CLINC150 benchmarks, respectively. Since the class data in ATIS and HWU64 has a long-tail distribution, we use the data of the top 10 and 50 frequent classes. The statistics of the four benchmarks are shown in Appendix \ref{appendix1}.
\subsection{Implementation Details}
At each step of the LID task, we report the accuracy on the testing data of all observed classes, denoted as $\text{acc}_i$. After the last step, we report Average Acc, which is the average accuracy of all step ($\frac {1} {K} \sum_{i=1}^{K} \text{acc}_i$), and Whole Acc, which is the accuracy on the whole testing data of all classes.
We use BERT in the HuggingFace's Transformers library. All hyper-parameters are obtained by a grid search on the validation set. The learning rate is $5e-5$ and the batch size is 64. The hyper-parameters $\tau$, $\alpha$, $\beta_1$,$\beta_2$, and $\beta_3$ are $50$, $-0.1$, $0.001$, $0.002$, and $10000$. $T=2$ in our method. The memory size is 200.
\subsection{Baselines}
In this work, we propose a model-agnostic lifelong learning method to handle the LID task. Therefore, we adopt other model-agnostic lifelong learning methods that achieve state-of-the-art performance on other tasks as our baselines. \textbf{EWC} \cite{emr} adopts an $L_2$ loss to slow down the update of important parameters. \textbf{LwF} \cite{lwf} uses knowledge distillation to learn the soft labels of the last model. \textbf{EMR} \cite{emr} randomly stores some old samples. \textbf{iCaRL} \cite{icarl} combines knowledge distillation and prototype-based sample selection in their method. \textbf{EEIL} \cite{EEIL} handles the data imbalance problem by resampling a balanced subset. \textbf{EMAR} \cite{emar} uses K-Means to select samples and consolidates the model by old prototypes. \textbf{FineTune} directly fine-tunes the pre-trained model on new data. \textbf{UpperBound} use training data of all observed classes to train the model, which is regarded as the upper bound.
\subsection{Main Results}
Figure \ref{fig4} shows the accuracy ($\text{acc}_i$) during the whole lifelong learning process. We also list Average Acc and Whole Acc after the last step in Appendix \ref{appendix2}. From the results, we can see that: (1) our MSR achieves state-of-the-art performance, significantly outperforming the baselines by 2.27\%, 1.68\%, 3.16\%, and 3.57\% whole accuracy on the ATIS, SNIPS, HWU64, CLINC150 benchmarks, respectively. These baselines either ignore the data imbalance problem or handle it by a simple resampling approach, which leads to catastrophic forgetting. (2) compared to EMAR, our method saves computation time because our method is more refined. (3) There is still a gap between our method and the upper bound. It indicates that there remain some challenges to be addressed.

\subsection{Ablation Study}
In this section, we perform ablation studies on the proposed three components. The results are shown in Appendix \ref{appendix3}. Removing any component brings a performance degradation. It shows that our method can alleviate catastrophic forgetting through multi-strategy rebalancing, which addresses multiple adverse effects caused by the data imbalance problem. 
\section{Conclusion}
In this paper, we propose the lifelong intent detection task to handle continually emerging user intents. In addition, we propose multi-strategy rebalancing to address multiple adverse effects caused by the data imbalance problem. Experimental results on four constructed benchmarks demonstrate the effectiveness of our method.
\bibliographystyle{ACM-Reference-Format}
\bibliography{acmart}

\appendix
\section{Statistics of benchmarks}
\label{appendix1}
In this section, we show the statistics of the four constructed benchmarks in Table \ref{t0}.
\begin{table}[h]
	\centering
	\smallskip
	\caption{Statistics of the ATIS, SNIPS, HWU64, and CLINC150 benchmarks. ``Training'' is the number of training samples.}
	\resizebox{0.46\textwidth}{!}{
		\begin{tabular}{lccccc}
			\toprule
			Benchmark  & Training      & Validation & Test     & Classes & Steps\\ \midrule 
			ATIS       & 4384          & 490        & 817      & 10    & 10  \\
			SNIPS      & 13084         & 700        & 700      & 7     & 7  \\
			HWU64      & 14465         & 4827       & 4845     & 50    & 10  \\
			CLINC150   & 15000         & 3000       & 3000     & 150   & 10  \\
			\bottomrule 
		\end{tabular}
	}
	\label{t0}
\end{table}

\section{Results on the four benchmarks}
\label{appendix2}
In this section, we list the results after the last step in Table \ref{t1}. The average accuracy of all steps and the whole accuracy of the whole testing data are shown in different columns. In both metrics, our method MSR significantly outperforms the baselines and achieves state-of-the-art performance on the four benchmarks. It implies that our method is effective in handling the LID task via multi-strategy rebalancing.
\begin{table*}[h]
	\centering
	\smallskip
	\caption{Average Acc and Whole Acc after the last step.}
	\resizebox{0.95\textwidth}{!}{
		\begin{tabular}{lllllllll}
			\toprule
			\multirow{2}{*}{Method} & \multicolumn{2}{c}{ATIS} & \multicolumn{2}{c}{SNIPS} & \multicolumn{2}{c}{HWU64} & \multicolumn{2}{c}{CLINC150} \\\cline{2-3}\cline{4-5}\cline{6-7}\cline{8-9}
			& Average Acc            & Whole Acc         & Average Acc        & Whole Acc   & Average Acc        & Whole Acc   & Average Acc        & Whole Acc      \\ \midrule
			FineTune   & 83.91  & 77.48  & 38.37  & 17.71  & 19.49 & 2.72  & 30.15  & 10.37         \\
			UpperBound & 99.78  & 99.27  & 99.27  & 97.71  & 71.57 & 68.34 & 97.25  & 95.63\\ \midrule 
			LwF        & 85.28  & 79.12  & 70.23  & 33.86  & 24.30 & 8.72  & 40.57  & 22.73        \\
			EWC        & 87.97  & 81.76  & 80.84  & 47.57  & 29.92 & 11.66 & 54.33  & 31.03        \\
			EMR        & 96.83  & 94.55  & 96.07  & 88.29  & 56.38 & 45.97 & 85.12  & 71.30        \\ 
			iCaRL      & 97.07  & 95.23  & 94.31  & 85.57  & 56.98 & 46.54 & 85.27  & 73.47       \\ 
			EEIL       & 97.50  & 95.42  & 95.26  & 85.86  & 58.63 & 48.98 & 86.74  & 74.43       \\ 
			EMAR       & 97.87  & 95.53  & 96.93  & 91.89  & 56.28 & 44.69 & 85.14  & 72.80       \\ 
			\textbf{MSR} (\textbf{Ours})& \textbf{99.03} & \textbf{97.80} & \textbf{97.64}      & \textbf{93.57} & \textbf{60.81} & \textbf{52.14} & \textbf{89.53}      & \textbf{78.00}
			\\ \bottomrule
		\end{tabular}
	}
	\label{t1}
\end{table*}
\begin{table*}[h]
	\centering
	\smallskip
	\caption{Ablation studies of multi-strategy rebalancing. We compare MSR with variants employing different components.}
	\resizebox{0.95\textwidth}{!}{
		\begin{tabular}{lllllllll}
			\toprule
			\multirow{2}{*}{Method} & \multicolumn{2}{c}{ATIS} & \multicolumn{2}{c}{SNIPS} & \multicolumn{2}{c}{HWU64} & \multicolumn{2}{c}{CLINC150} \\\cline{2-3}\cline{4-5}\cline{6-7}\cline{8-9}
			& Average Acc            & Whole Acc         & Average Acc        & Whole Acc   & Average Acc        & Whole Acc   & Average Acc        & Whole Acc      \\ \midrule
			\textbf{MSR} (\textbf{Ours}) & \textbf{99.03} & \textbf{97.80} & \textbf{97.64}      & \textbf{93.57} & \textbf{60.81} & \textbf{52.14} & \textbf{89.53}      & \textbf{78.00}\\
			- CN         & 98.88 & 96.94 & 97.54 & 93.36 & 60.34 & 51.95 & 89.46 & 77.10        \\
		    - FKD        & 98.31 & 96.21 & 97.51 & 93.14 & 59.75 & 50.03 & 89.41 & 76.77 \\
			- PKD        & 98.61 & 96.82 & 97.31 & 92.57 & 59.61 & 51.02 & 89.08 & 75.70        \\
			- HKD        & 98.23 & 95.84 & 96.63 & 92.00 & 59.60 & 49.14 & 88.54 & 74.27        \\
			- ICML       & 98.52 & 96.70 & 97.04 & 92.29 & 59.56 & 48.24 & 89.26 & 76.90       \\
			- CN and HKD & 97.79 & 95.23 & 96.29 & 91.43 & 58.97 & 47.78 & 87.34 & 72.23       \\ 
			- MSR        & 96.83 & 94.55 & 96.07 & 88.29 & 56.38 & 45.97 & 85.12 & 71.30      \\ \bottomrule
		\end{tabular}
	}
	\label{t2}
\end{table*}
\section{Ablation Study}
\label{appendix3}
Our method consists of three components: cosine normalization, hierarchical knowledge distillation, and inter-class margin loss. We show the ablation studies of the three components. The results are shown in Table \ref{t2}. For ``- CN'', we replace cosine normalization with the dot product similarity. For ``- FKD'', we remove the feature-level knowledge distillation. For ``- PKD'', the prediction-level knowledge distillation is removed. For ``- HKD'', this model does not adopt the proposed hierarchical knowledge distillation. For ``- ICML'', the model removes the inter-class margin loss. For ``- CN and HKD'', we remove both cosine normalization and hierarchical knowledge distillation. The model without multi-strategy rebalancing (``- MSR'', i.e., the model EMR) is shown in the last row. We can see that these variants achieve low performance. It indicates that simultaneously utilizing these multiple strategies is very effective.

\end{document}